\documentclass[times, review, 10pt]{elsarticle}

\usepackage{amssymb}
\usepackage{amsthm}
\usepackage{amsmath}
\usepackage{derivative}
\usepackage{multirow}
\usepackage{booktabs}
\usepackage{comment}
\usepackage{lineno}
\usepackage{adjustbox}

\DeclareMathOperator*{\argmax}{arg\,max}

\usepackage{color}
\newcommand{\rr}{\color{red}}
\newcommand{\bb}{\color{black}}

\usepackage{setspace}

\usepackage[utf8]{inputenc}
\usepackage[T1]{fontenc}
\usepackage[swedish,english]{babel}

\journal{Pattern Recognition}

\begin{document}

\begin{frontmatter}



\title{Multimodal Explainability via Latent Shift applied to COVID-19 stratification}

\author[campus]{Valerio Guarrasi}
\author[campus]{Lorenzo Tronchin}
\author[galeazzi,milano]{Domenico Albano}
\author[santanna]{Eliodoro Faiella}
\author[cdi]{Deborah Fazzini}
\author[campus,santanna]{Domiziana Santucci}
\author[campus,umea]{Paolo Soda\corref{correspondingauthor}}
\ead{paolo.soda@umu.se}
\cortext[correspondingauthor]{Corresponding author}

\affiliation[campus]{organization={Unit of Computer Systems and Bioinformatics, Department of Engineering, University Campus Bio-Medico of Rome}, 
city={Rome},
country={Italy}}
\affiliation[galeazzi]{organization={IRCCS Istituto Ortopedico Galeazzi},
city={Milan},
country={Italy}}
\affiliation[milano]{organization={Department of Biomedical, Surgical and Dental Sciences, Università degli Studi di Milano},
city={Milan},
country={Italy}}
\affiliation[santanna]{organization={Department of Radiology, Sant'Anna Hospital},
city={San Fermo della Battaglia, Como},
country={Italy}}
\affiliation[cdi]{organization={Department of Diagnostic Imaging and Stereotactic Radiosurgey, Centro Diagnostico Italiano S.p.A.},
city={Milan},
country={Italy}}
\affiliation[umea]{organization={Department of Diagnostics and Intervention, Radiation Physics, Biomedical Engineering, Umeå University},
city={Umeå},
country={Sweden}}

\begin{abstract}

We are witnessing a widespread adoption of artificial intelligence in healthcare. 
However, most of the advancements in deep learning in this area consider only unimodal data, neglecting other modalities. 
Their multimodal interpretation necessary for supporting diagnosis, prognosis and treatment decisions.
In this work we present a deep architecture, which jointly learns modality reconstructions and sample classifications using tabular and imaging data.
The explanation of the decision taken is computed by applying a latent shift that, simulates a counterfactual prediction revealing the features of each modality that contribute the most to the decision and a quantitative score indicating the modality importance.
We validate our approach in the context of COVID-19 pandemic using the AIforCOVID dataset, which contains multimodal data for the early identification of patients at risk of severe outcome.
The results show that the proposed method provides meaningful explanations without degrading the classification performance.

\end{abstract}




\end{frontmatter}


\section{Introduction} \label{sec:introduction}

In the last decade, the practice of modern medicine has started to heavily rely on the utilization of data coming from multiple sources~\cite{bib:huang2020fusion}.
At the same time, artificial intelligence (AI) has achieved state-of-the-art results in various domains~\cite{bib:caruso2022multimodal}, healthcare included.
Nevertheless, most of the deep neural networks applied to medical tasks consider only one data modality, neglecting information available in other sources.
However, analyzing medical findings is multimodal by its very nature: a characteristic that, in turn, asks for developing AI approaches able to process data of different modalities~\cite{bib:cipollari2022convolutional,bib:baltruvsaitis2018multimodal}.
This has fostered the rise of multimodal deep learning (MDL)~\cite{bib:baltruvsaitis2018multimodal}, which aims to develop learning models able to process and link information gathered from different modalities.
MDL is a topic where researchers have investigated several methods to learn together multimodal information via early, late, and joint fusion~\cite{bib:baltruvsaitis2018multimodal}, as it will be presented in section~\ref{sec:background}.


With the goal of reaching high performance, many complex AI models developed so far have a black-box nature~\cite{bib:tjoa2020survey}, neglecting trust and transparency~\cite{bib:arrieta2020explainable}, two features that are of particular importance in biomedicine~\cite{bib:ramachandram2017deep}.
Indeed, a lack in explainability limits the application of AI models into the clinical practice.
To overcome this limitation, in the last years large research efforts have been directed towards explainable AI (XAI), which aims to explain how a black-box model produces its outcomes.

In healthcare, the need of multimodal models and of explainability make multimodal explanations vital to develop robust and trustworthy AI models.
This happens because multimodal models extract more comprehensive information than the unimodal models, so that their explanations could offer more insights into the available medical data.
The multimodal setting of XAI explores the complementary and explanatory strengths of the different modalities, with the goal of obtaining better explanations that localize the relevant features and modalities~\cite{bib:park2018multimodal}. 
Despite this relevance, to the best of our knowledge, the biomedical literature lacks of explainable deep multimodal models.
For this reason we present a novel end-to-end multimodal architecture, with intrinsic explanations, that jointly learns modality reconstructions and multimodal classification using imaging and tabular modalities.
With this architecture we extract deep multimodal representations of the data; then, we apply a latent shift to simulate a counterfactual prediction, thus obtaining an intrinsic explanation that reveals how and why the model arrived at a particular decision.

We test our approach in the context of the COVID-19 pandemic using the AIforCOVID public dataset~\cite{bib:soda2021aiforcovid} for three reasons.
First, the development of AI-based tools supporting COVID-19 prognosis exploiting multimodal data is still an open research issue addressed by few work in the literature~\cite{bib:soda2021aiforcovid,bib:signoroni2021bs,bib:zhu2020deep,bib:al2020classifier, bib:bai2020predicting,bib:ning2020open,bib:fang2021deep}, which will be discussed in section~\ref{sec:background}.
Second, while there is a lack of multimodal XAI (MXAI) approaches in general, no one has proposed multimodal explanations to gain trust and transparency in COVID-19 prognosis.
Third, the AIforCOVID dataset is the largest publicly available repository containing chest X-ray (CXR) images and clinical data collected at the time of hospitalization, further to clinical outcomes stratifying patients into those with and without  risk of a severe disease progression~\cite{bib:santa2021public}.
It is worth noting that, in general, images and clinical data are two important sources of information in medicine.
Indeed, the former allows radiologists to focus on visual evidence for both diagnostic and prognostic purposes, whereas the latter which are usually stores as tabular data, offer to clinicians a concise and multi-dimensional assessment of patients' health status.
Hence, having both the modalities should be important to test MDL and MXAI approaches that would support the medical decision process.


The main contributions of our work are:
\begin{itemize}
    \item The development of an intrinsic explainable architecture specifically designed for multimodal classification.
    
    \item The introduction of a joint learning approach that enables to simultaneously train both  the data reconstruction and the classification tasks using tabular and imaging data.
    
    \item The proposal of a novel latent space counterfactual method that allows for explainability in both multimodal and unimodal contexts. It reveals the modalities and features that contribute the most to the decision-making process.
   
    \item The effective application of the proposed approach in the context of the COVID-19 pandemic to early identify patients at risk of severe outcomes, using the publicly available AIforCOVID dataset.
   
    \item The  validation of the proposed method, strengthened by a reader study with four radiologists, showing  that it provides meaningful explanations without sacrificing the classification performance.
   
   
\end{itemize}

The rest of the manuscript is organized as follows.
Section~\ref{sec:background} introduces the state-of-the art of both MDL and MXAI.
Then, section~\ref{sec:methods} presents our novel architecture and training procedure, and it explains the MXAI method extracting multimodal explanations.
In section~\ref{sec:experimental} we describe the dataset used to validate the methods, the pre-processing phase on the data, the implementation setup and the validation strategy adopted.
Section~\ref{sec:results} presents and discusses the obtained results, whilst section~\ref{sec:conclusion} provides concluding remarks.

\section{Background}  \label{sec:background}

In this section we first present the state-of-the-art of MDL in the context of COVID-19 prognosis prediction by using images and tabular clinical data, whereas the interested readers can deepen  MDL in healthcare in recent surveys~\cite{bib:huang2020fusion}.
Indeed the literature is quite large and its review is out of the scope of this work.
Second, given the lack of MXAI approaches in COVID-19 prognosis~\cite{bib:fiscon2021assessing}, we will summarize MXAI research in healthcare, whilst the readers can refer to~\cite{bib:abiodun2022explainable} for XAI applications on unimodal data in COVID-19 imaging.
We will conclude this section by summarizing the motivations of this work.

\subsection{MDL}

There is a general consensus that medical images complemented by clinical data can help physicians, and radiologist in particular, better understanding the patient's state, thus advancing to a more informative decision making process~\cite{bib:boonn2009radiologist}.
In this respect, research in the field of MDL has been increasing~\cite{bib:baltruvsaitis2018multimodal} since multimodal data give the opportunity to train models that can learn the complex dynamics behind a disease.

The level at which the fusion of input modalities occurs in the network is usually distinguished into early, intermediate or late fusion~\cite{bib:baltruvsaitis2018multimodal}.
Early fusion combines raw data or extracted features by the different modalities which are fed to a simple learner, whilst late fusion combines at the decision level the outputs of networks independently trained on each modality.
Intermediate fusion learns a joint representation of different modalities at a shared representation layer, by propagating the loss back to the feature extractor network in an end-to-end manner.

The application of AI in COVID-19 using medical imaging and clinical data has mainly focused on discriminating patients suffering from COVID-19 pneumonia from those which are healthy or affected by different types of pneumonia~\cite{bib:wynants2020prediction}.
Nevertheless, only four work   investigated   patients' stratification into mild and severe outcomes using such multimodal data~\cite{bib:soda2021aiforcovid,bib:bai2020predicting,bib:ning2020open,bib:fang2021deep}, which can be further divided in three using computed tomography (CT) scans~\cite{ bib:bai2020predicting,bib:ning2020open,bib:fang2021deep} and one using CXR images~\cite{bib:soda2021aiforcovid}.

In~\cite{bib:bai2020predicting} the authors proposed a deep network using CT scans and $53$ clinical features to detect the potential malignant progression of mild patients.
Via a multi-layer perceptron (MLP) processing  the clinical features only, their method gets an embedding that is then concatenated in an early fusion approach with the flattened CT scan.
This new vector  then fed  a Long short-term memory (LSTM) network followed by a fully-connected (FC) network, which produces the output.
On a private cohort of 199 patients, they obtained an accuracy of $79.20\%$.
In~\cite{bib:ning2020open} the authors used $130$ clinical features and CT scans to discriminate between negative, mild and severe COVID-19 cases via an early fusion of a VGG-16 and a 7-layer FC network. 
They used a private dataset containing 1521 patients, achieving an accuracy equal to $81.10\%$.
Fang et al applied joint fusion combining  CT scans and $61$ clinical features, which fed a deep network to predict COVID-19 malignant progression~\cite{bib:fang2021deep}. 
The approach  extracts  the abstract representations of   CT images using a 3D ResNet, and of the  clinical data via a FC network.
These two embeddings are concatenated and given to an LSTM followed by another  FC network.
The whole architecture is training in an end-to-end modality. 
On a private dataset containing 1040 patients they achieved  an accuracy equal to  $87.70\%$.

It is worth noting that CXR helps indicating abnormal formations of a large variety of chest diseases by using a very small amount of radiation, whilst CT delivers a much higher detail level of the lungs' structures.
Furthermore, X-ray equipment is much smaller, less complex, and with lower costs than CT scans;  it also prevents other three limitations of CT imaging, i.e., the   lack of available machines' slots, the difficulty of moving bedridden patients, and the long sanitation times.
For these reasons, several authors indicated that CXR imaging  fit well with the needs of  COVID-19 pandemic~\cite{bib:roberts2021common}.
Let us now focus on the only work that uses CXR images  and clinical data for the COVID-19 prognosis~\cite{bib:soda2021aiforcovid}.
There, the authors presented three different multimodal AI approaches, offering also   baseline performance on the AIforCOVID dataset, which we will present in section~\ref{sec:experimental} together with the best results attained.
The first method, named as handcrafted approach, computed  first-order and texture features from the images, which are then stacked in an early fusion fashion with the clinical features, then feeding different learners among which the Support Vector Machine resulted to be the best.
The second approach, referred to as hybrid, combines  automatic features extracted  by a convolutional  neural network with the clinical ones; then it runs a feature selection stage whose output is given to the learner. 
The hybrid approach achieved the best results using the GoogleNet and the Support Vector Machine classifier. 
The third approach, named as end-to-end, performs intermediate fusion of the two modalities by defining  a multi-input network concatenating hidden vectors of the two modalities.
This architecture contains three main branches: two process independently CXR scans and clinical features  to get  a small number of relevant and abstract features, while the third one  concatenates such embeddings that is given to a FC network, which outputs the prognosis.

\subsection{MXAI}

High performing deep models are often black-boxes, which hide their decision-making process, making it hard to understand why a certain result is obtained.
This has boosted the growth of XAI, and many unimodal methods have been proposed to extract explanations on how the model has interpreted the data~\cite{bib:arrieta2020explainable} with applications to different fields.
In particular, many authors agree that  explanations are strongly recommended in medical applications~\cite{bib:guidotti2018survey}, because this would help mapping explainability with causability that, in turns, would allow  practitioners to understand why a model came up with a result.

The only available review on MXAI~\cite{bib:joshi2021review} surveys its applications in computer vision and natural language processing, showing that MXAI lacks in the medical field.
This is confirmed by the position paper~\cite{bib:holzinger2020explainable}, which states that in radiology there is a lack of integrative methods that, combining imaging and tabular data, provide explanations on the decisions taken.
This confirms the need of multimodal explanations to capture the complexity of all the factors underlying a disease.
Indeed, for a medical task to have a comprehensive global view of the data and of the system, an ideal MXAI method should be able to identify the importance of each modality and the importance of each unimodal feature.

The MXAI review~\cite{bib:joshi2021review}, even if it does not focus on the medical field, is also interesting because it groups XAI algorithms adopting three different criteria.
First, it focuses on the stage at which the XAI can be applied, identifying pre-modeling, during modelling and post-hoc modeling  explanations~\cite{bib:joshi2021review}.
As their names explain, the pre-modeling methods' explainability is included before the model development, during modelling include the models which are usually explainable by design and employ intrinsic methods, and post-hoc modelling is applied after the model is developed by extracting explanations via perturbations or backpropagation methods~\cite{bib:selvaraju2016grad}.
Second, with reference to the scope of the explanation, XAI models can be either local or global~\cite{bib:joshi2021review}, depending if the explanation regards a single instance or the model as a whole.
The third criterion deal with the dependency of the XAI algorithm, so that it distinguishes model specific and model agnostic explanations~\cite{bib:joshi2021review}.
While the interested readers can deepen~\cite{bib:joshi2021review} to have more details, on the basis of this survey we observe that there is a lack in multimodal intrinsic explainability, i.e., methods able to return multimodal local explanations.

As mentioned in the forewords of this section, the analysis of the literature that uses multimodal data for COVID-19 prognosis shows that, to the best of our knowledge, none has investigated MXAI in this field yet.

\subsection{Motivations}

Multimodal settings have improved the predictive power of models in many applications thanks to the interaction of different modalities, via a richer representation with task-relevant features~\cite{bib:park2018multimodal}.
Nevertheless, this availability of  information from different  modality makes  explainability a key necessity to reduce the opacity of the multimodal deep architectures~\cite{bib:ramachandram2017deep}.
This has recently fostered  the raise of MXAI, which   has mainly focused on computer vision and natural language processing. 
Indeed, the   literature on XAI in medical applications has concentrated more on unimodal attribution methods, struggling in having explanations of neural networks working on multiple data sources. 
Therefore,  developing multimodal methods for explainability is an urgent and open issue, also because the   development of   multimodal deep architectures in different healthcare applications  asks for novel approaches to open such black boxes. 
In turn, this can help physicians, patients and regulators to trust the decisions taken. 
Among the several fields where MDL and MXAI can be applied, we test our methodology to the early identification of COVID-19 patients at risk of severe outcome using imaging and tabular data,  because the survey of the literature presented hereinbefore shows that few work has addressed  this challenge, despite the disruptive impact of this disease worldwide.

\section{Methods} \label{sec:methods}

In this section we present a novel architecture that exploits joint learning, for which we design an intrinsic counterfactual MXAI approach to extract explanations of a classification task.
In general, counterfactual explanations refer to a type of explanation that aims to comprehend the causes of an observed outcome by exploring alternative scenarios, which helps in gaining a deeper understanding of the causal relationships that led to the observed outcome~\cite{bib:tjoa2020survey}.
Such multimodal explanations will permit users to understand not only the importance of each modality for each classification, but also the features which contributed the most to the decision for every single modality.

We first present the architecture of the multimodal model; second we focus on the training approach and, third, we detail the intrinsic MXAI method.

\subsection{Notation}

The notation used henceforth makes us of the following symbols:
\begin{itemize}
    \item $T$ and $I$ are the tabular and the imaging modalities, respectively;
    \item $\boldsymbol{x}_T$ and $\boldsymbol{x}_I$ are the inputs for the tabular and imaging modality, respectively;
    \item $AE$ and $CAE$ are the autoencoder and convolution autoencoder which receive as input $\boldsymbol{x}_T$ and $\boldsymbol{x}_I$, respectively. Both are composed of an encoder $E_{AE}$, $E_{CAE}$ and a decoder $D_{AE}$, $D_{CAE}$, respectively;
    \item $\boldsymbol{h}_T \in \mathbb{R}^n$ and $\boldsymbol{h}_I \in \mathbb{R}^m$ are the latent vectors of the $AE$ and the $CAE$, respectively. Their concatenation produces the multimodal embedding $\boldsymbol{h} \in \mathbb{R}^{n+m}$;
    \item $\hat{\boldsymbol{x}}_T$ and $\hat{\boldsymbol{x}}_I$ are the outputs produced by the $AE$ and the $CAE$, respectively, representing the reconstruction of the inputs $\boldsymbol{x}_T$ and $\boldsymbol{x}_I$;
    \item $C_{MLP}$ is the multi-layer perceptron receiving the vector $\boldsymbol{h}$ to perform the classification;
    \item $\boldsymbol{y} \in \mathbb{R}^c$ is the output vector of $C_{MLP}$, which expresses the predicted posterior probability, with $c$ being equal to the number of classes;
    \item $L_T$, $L_I$, $L_C$ are the loss functions of the $AE$, $CAE$ and $C_{MLP}$, respectively, whose linear combination results in $L$, weighted by the corresponding scalar parameters $\gamma_T \in \mathbb{R}$, $\gamma_I \in \mathbb{R}$, $\gamma_C \in \mathbb{R}$;
    \item $\boldsymbol{h}_T^{\lambda} \in \mathbb{R}^n$, $\boldsymbol{h}_I^{\lambda} \in \mathbb{R}^m$ and $\boldsymbol{h}^{\lambda} \in \mathbb{R}^{n+m}$ are the modified vector embeddings of $\boldsymbol{h}_T$, $\boldsymbol{h}_I$ and $\boldsymbol{h}$, respectively, regulated by scalar parameter $\lambda \in \mathbb{R}$; 
    \item $\hat{\boldsymbol{x}}_T^{\lambda}$, $\hat{\boldsymbol{x}}_I^{\lambda}$ and $\boldsymbol{y}^{\lambda} \in \mathbb{R}^c$ are the outputs produced by $D_{AE}$, $D_{CAE}$ and $C_{MLP}$, respectively, when the input is $\boldsymbol{h}_T^{\lambda}$, $\boldsymbol{h}_I^{\lambda}$ and $\boldsymbol{h}^{\lambda}$, respectively;
    \item $\Delta_T \in \mathbb{R}$ and $\Delta_I \in \mathbb{R}$ express the resulting modality importance comparing $\boldsymbol{h}_T$ with $\boldsymbol{h}_T^{\lambda}$ and $\boldsymbol{h}_I$ with $\boldsymbol{h}_I^{\lambda}$, respectively;
    \item $\hat{\boldsymbol{\Delta}}_T \in \mathbb{R}^n$, $\hat{\boldsymbol{\Delta}}_I \in \mathbb{R}^m$ express the resulting unimodal feature importance comparing $\hat{\boldsymbol{x}}_T$ with $\hat{\boldsymbol{x}}_T^{\lambda}$ and $\hat{\boldsymbol{x}}_I$ with $\hat{\boldsymbol{x}}_I^{\lambda}$, respectively.
\end{itemize}

\subsection{Architecture}

\begin{figure}[t]
\centering
\caption{Schematic view of the multimodal deep architecture: for each instance, the input modalities $\boldsymbol{x}_T$ and $\boldsymbol{x}_I$ feed into their corresponding encoders $E_{AE}$ and $E_{CAE}$, obtaining the unimodal embeddings $\boldsymbol{h}_T$ and $\boldsymbol{h}_I$, respectively. These embeddings are then concatenated into the multimodal embedding $\boldsymbol{h}$, which subsequently feeds into the decoders $D_{AE}$, $D_{CAE}$, and the classifier $C_{MLP}$. The resulting outputs are the reconstructions $\hat{\boldsymbol{x}}_T$, $\hat{\boldsymbol{x}}_I$, and classification $\boldsymbol{y}$, respectively. The model is trained by simultaneously minimizing the reconstruction losses $L_{RT}$, $L_{RI}$, and the classification loss $L_C$.}
\includegraphics[width=\textwidth]{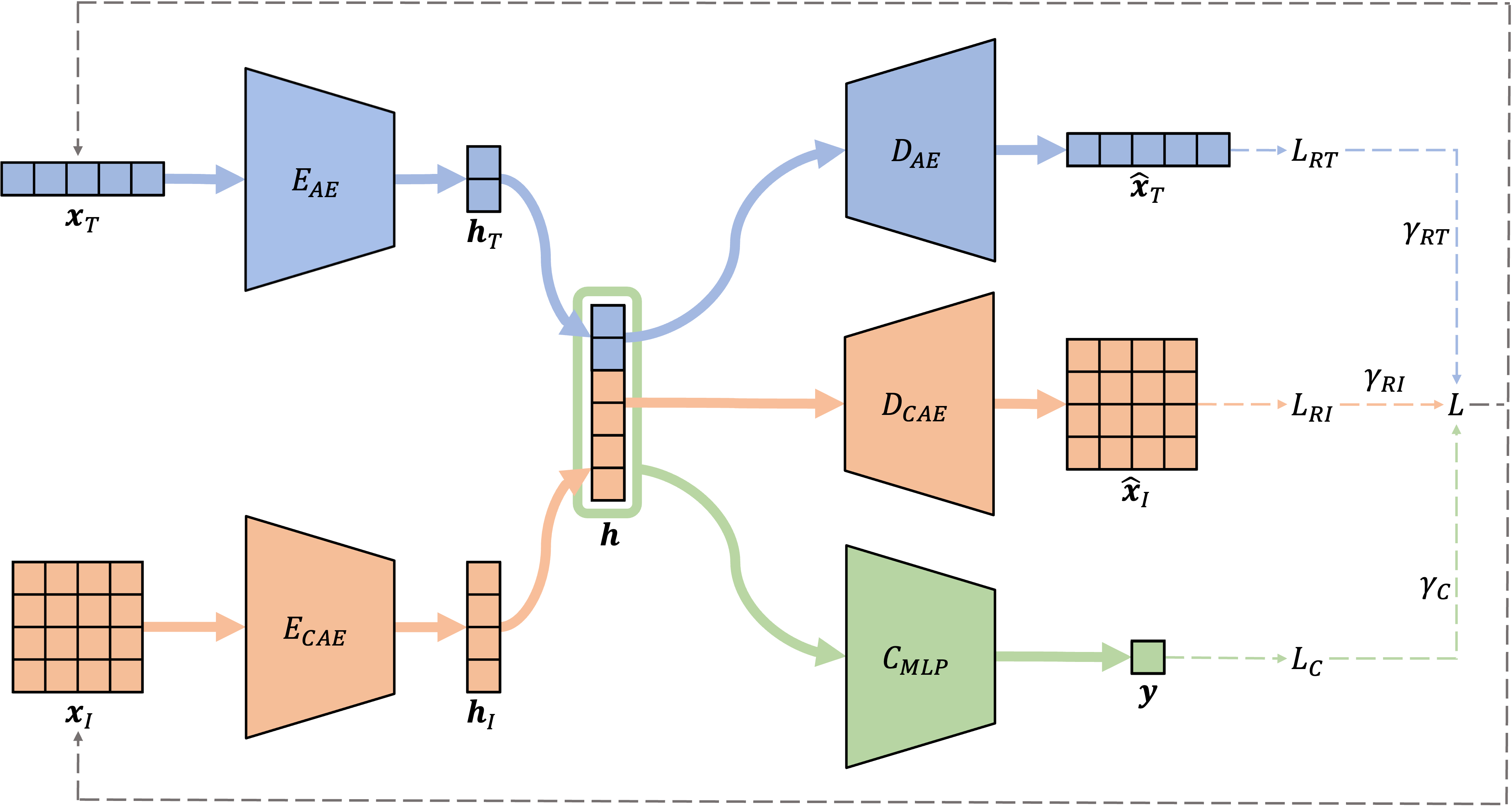}
\label{fig:architecture}
\end{figure}

Here we present the structure of the designed classification model that works with $T$ and $I$.
The proposed multimodal architecture consists of three blocks: an autoencoder ($AE$), a convolutional autoencoder ($CAE$), and a multi-layer perceptron classifier ($C_{MLP}$).
As shown in Figure~\ref{fig:architecture}, the network has two inputs and three outputs.
The tabular modality $\boldsymbol{x}_T$ feeds the $AE$, whereas the imaging modality $\boldsymbol{x}_I$ is given to the CAE.
Both are composed of an encoder and a decoder, returning the reconstruction of the respective modality, $\hat{\boldsymbol{x}}_T$ and $\hat{\boldsymbol{x}}_I$.
By concatenating the two embeddings $\boldsymbol{h}_T$ and $\boldsymbol{h}_I$ we get $\boldsymbol{h}$, which is given to the $C_{MLP}$ classifier that returns the classification vector $\boldsymbol{y}$.
The entire architecture is trained in an end-to-end manner, via a linear combination of three loss functions, two for reconstruction ($L_{T}$ and $L_{I}$) and one for the classification ($L_C$).

This overview reveals that our framework jointly learns deep representations of imaging and tabular data to perform a classification task. 
Indeed, it learns a feature space with local modality structure able to be used for reconstruction, and it manipulates the combined feature space by incorporating a classification oriented loss.

\paragraph{Autoencoders}

An $AE$ and a $CAE$ are artificial neural networks which learn an approximation to the identity function, with the goal of minimizing the distance between the outputs $\hat{\boldsymbol{x}}_T$, $\hat{\boldsymbol{x}}_I$ and the inputs $\boldsymbol{x}_T$, $\boldsymbol{x}_I$, respectively.
The encoders $E_{AE}$ and $E_{CAE}$ compress the corresponding inputs $\boldsymbol{x}_T$, $\boldsymbol{x}_I$ to a latent space representation $\boldsymbol{h}_T$ and $\boldsymbol{h}_I$, using fully connected layers in the $AE$ and convolutional layers in the $CAE$, respectively.
The decoders $D_{AE}$ and $D_{CAE}$ use the bottleneck latent space representation $\boldsymbol{h}_T$ and $\boldsymbol{h}_I$ to reconstruct the inputs $\boldsymbol{x}_T$, $\boldsymbol{x}_I$ in $\hat{\boldsymbol{x}}_T$, $\hat{\boldsymbol{x}}_I$, respectively.
Therefore:
\begin{equation}
    \boldsymbol{h}_T = E_{AE}(\boldsymbol{x}_T) \label{eq:h_clinical}
\end{equation}
\begin{equation}
    \boldsymbol{h}_I = E_{CAE}(\boldsymbol{x}_I) \label{eq:h_image}
\end{equation}
\begin{equation}
    \hat{\boldsymbol{x}}_T = D_{AE}(\boldsymbol{h}_T)
\end{equation}
\begin{equation}
    \hat{\boldsymbol{x}}_I = D_{CAE}(\boldsymbol{h}_I)
\end{equation}
When training the $AE$ and the $CAE$ we aim to minimize the distance between its inputs and outputs over all samples, using as reconstruction loss functions $L_{T}$ and $L_{I}$ for the tabular and imaging modalities, respectively.
We constrain the dimension of latent spaces $\boldsymbol{h}_T$ and $\boldsymbol{h}_I$ to be lower than input data $\boldsymbol{x}_T$ and $\boldsymbol{x}_I$, respectively, forcing both the $AE$ and the $CAE$ to capture the most salient features of the data.
This is a well-known approach to avoid identity mapping~\cite{bib:bahadur2022dimension}.

\paragraph{Classifier}

The two embeddings $\boldsymbol{h}_T$ and $\boldsymbol{h}_I$ are concatenated in $\boldsymbol{h}$ and used as input to the fully connected $C_{MLP}$, which performs the classification task returning $\boldsymbol{y}$,
So that:
\begin{equation}
    \boldsymbol{y} = C_{MLP}(\boldsymbol{h})
\end{equation}
The goal of this block is to minimize the classification error with a classification loss $L_C$.
Note that the final layer of the $C_{MLP}$ uses the Softmax activation function, such that
\begin{equation}
    \sum_{i=1}^{c} \boldsymbol{y}_i = 1
\end{equation}
This implies that $\boldsymbol{y}$ can be considered as an estimate of the posterior probability.

\paragraph{End-to-end training}

In this way, the network's training can be back-propagated in an end-to-end manner, via a linear combination of the three loss functions:
\begin{equation}
    L = \gamma_{T} L_{T} + \gamma_{I} L_{I} + \gamma_{C} L_C
\end{equation}
where $\gamma_{T}$, $\gamma_{I}$, and $\gamma_{C}$ are parameters, which regulate the importance of each loss function.

This approach has the beneficial effect of being able to learn embedded features in an end-to-end way, which are jointly used to perform data reconstruction and classification, minimizing the reconstruction loss of $AE$ and the $CAE$ and the classification loss of the $C_{MLP}$.
Our key idea is that the co-learning of the $AE$, the $CAE$ and the $C_{MLP}$ is beneficial to learn features from the tabular and imaging modality to obtain a classification and a good reconstruction useful for the explainability, presented in section~\ref{subsec:MXAI}.

\subsection{Three-stage training} \label{subsec:three_stage}

Given the complex structure of the architecture proposed, we train the network with a three-stage procedure, which adapts the $\gamma_{T}$, $\gamma_{I}$, $\gamma_{C}$ parameters in way to concentrate the training on different parts of the network.
The three stages are:
\begin{enumerate}
    \item Setting $\gamma_{T}=1$, $\gamma_{I}=0$ and $\gamma_{C}=0$ to train only the weights of the $AE$;
    \item Setting $\gamma_{T}=0$, $\gamma_{I}=1$ and $\gamma_{C}=0$ to train only the weights of the $CAE$;
    \item Setting $\gamma_{T}=1$, $\gamma_{I}=1$ and $\gamma_{C}=1$ to train all the weights of network.
\end{enumerate}

The main idea is to help the training of the $C_{MLP}$ classifier, giving initialization weights that constrict an optimal modality embedding for reconstruction, expressing a good summary of the data.
Notice also that, given the architecture of the network, it is irrelevant if we invert stage 1 and 2 since the $AE$ and $CAE$ have no weights in common.
In step 3 we decided to set all the parameters to $1$ so that all the tasks would have equal weight.

\subsection{MXAI} \label{subsec:MXAI}

We use a gradient update, also referred to as latent shift, that can transform the latent representation of the inputs to exaggerate or curtail the features used for prediction.
Via the latent shift explanations we obtain both modality importance and feature importance for each prediction.

\paragraph{Latent shift}

\begin{figure}[t]
\centering
\caption{Schematic view of the MXAI framework: once the model is trained, each instance's multimodal embedding $\boldsymbol{h}$ feeds into the decoders $D_{AE}$, $D_{CAE}$, and the classifier $C_{MLP}$, according to colors that specify that portion of $h$ is given to each network. The decoders and the classifier provide the original reconstructions $\hat{\boldsymbol{x}}_T$, $\hat{\boldsymbol{x}}_I$, and classification $\boldsymbol{y}$. Via the latent-shift method we obtain a $\lambda > 0$, which gives us a flip in the classification $\boldsymbol{y}^{\lambda}$ by feeding the shifted multimodal embedding $\boldsymbol{h}^{\lambda}$ to $C_{MLP}$. By feeding this new embedding to $D_{AE}$ and $D_{CAE}$, we obtain new reconstructions $\hat{\boldsymbol{x}}^{\lambda}_T$ and $\hat{\boldsymbol{x}}^{\lambda}_I$. By comparing $\boldsymbol{h}$ with $\boldsymbol{h}^{\lambda}$, $\hat{\boldsymbol{x}}_T$ with $\hat{\boldsymbol{x}}^{\lambda}_T$, and $\hat{\boldsymbol{x}}_I$ with $\hat{\boldsymbol{x}}^{\lambda}_I$, we obtain the corresponding multimodal and unimodal explanations, respectively.}
\includegraphics[width=\textwidth]{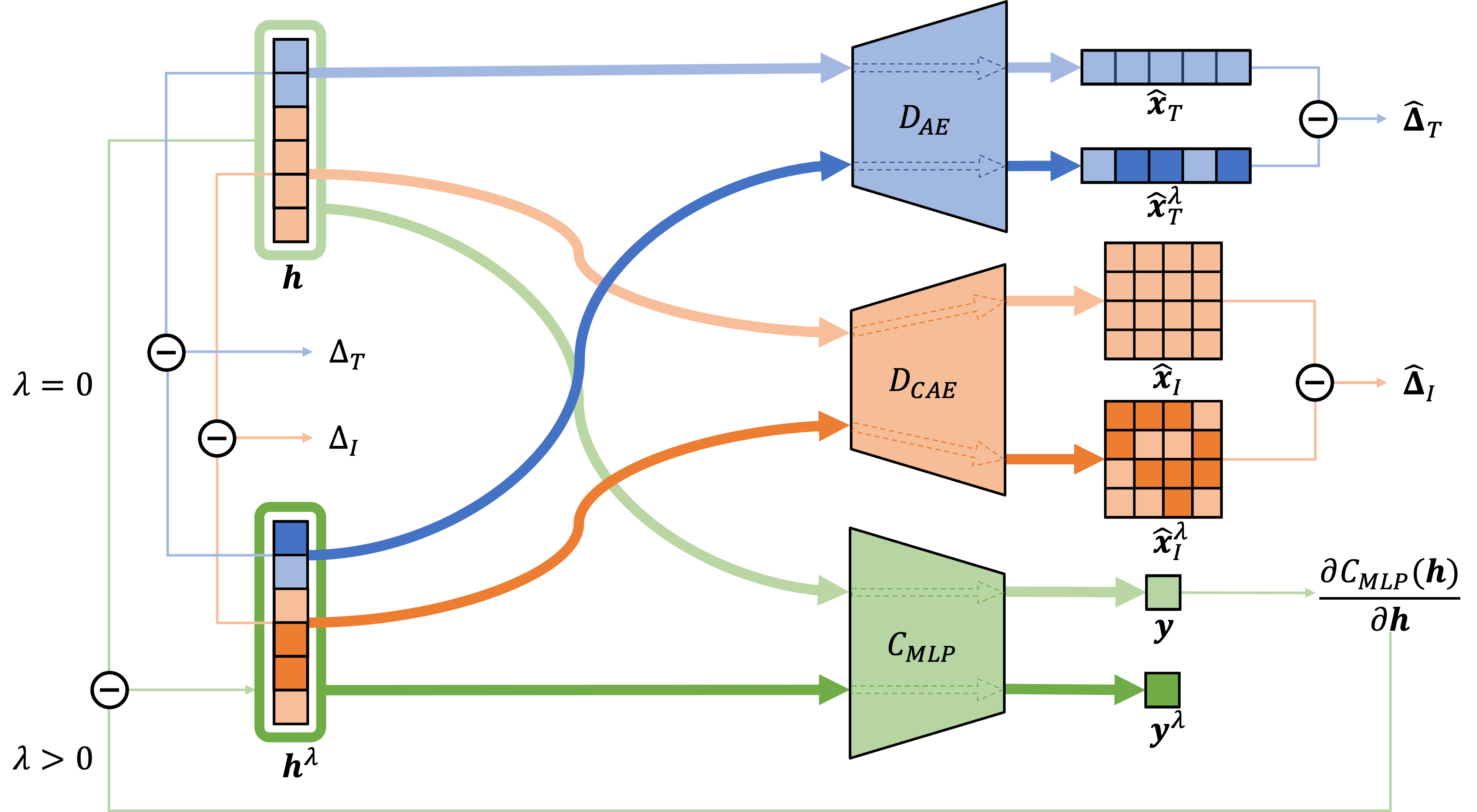}
\label{fig:xai}
\end{figure}

The only requisite to apply latent shift to a network is of having all the network components, which receive the latent vector $\boldsymbol{h}$, to be differentiable. 
With the $AE$, a $CAE$ and a $C_{MLP}$ we satisfy this requisite and, in addition, they are simple to implement and train.
Once these components are trained, we extract the explanation as shown in Figure~\ref{fig:xai}. 
Multimodal input instances $\boldsymbol{x}_T$ and $\boldsymbol{x}_I$ are encoded producing the multimodal latent representations $\boldsymbol{h}_T$ and $\boldsymbol{h}_I$, which are combined into $\boldsymbol{h}$, as already described.
Perturbations of this latent embedding are computed via
\begin{equation}
    \boldsymbol{h}^{\lambda} = \boldsymbol{h} - \lambda \pdv{C_{MLP}(\boldsymbol{h})}{\boldsymbol{h}}
\end{equation}
where $\lambda \in \mathbb{R}$ is a parameter  establishing how much the original embedding is modified.
With $\lambda > 0$, we expect that $C_{MLP}(\boldsymbol{h}^{\lambda})$ would provide a prediction $\boldsymbol{y}^{\lambda}$ so that
\begin{equation}
    \max({\boldsymbol{y}}) \geq {\boldsymbol{y}^{\lambda}(\argmax(\boldsymbol{y}))}
\end{equation}
This implies that, as $\lambda$ increases, we expect a flip of the predicted class.
In other words, guided by the direction of variation of the output in the latent space determined by the gradient of network output,  we are interested in determining the $\lambda$ value for which a classification label flip occurs.
With too small values of $\lambda$, for the smoothness principle, the difference between the original modality input and the reconstruction will not be  large enough to change the prediction of the model.
On the contrary, too large values of $\lambda$ would distort the reconstruction so much that it will not be useful for explainability.
Thus, to find the value of $\lambda$ where the class flip occurs, we use an iterative search that, starting from $\lambda= 0$, and using a  fixed step heuristically set to $10$, increases $\lambda$ until $\boldsymbol{y}^{\lambda} \neq \boldsymbol{y}$. 

We produce $\lambda$-shifted counterfactual multimodal reconstructions $\boldsymbol{x}_T^{\lambda}$, $\boldsymbol{x}_I^{\lambda}$ and output probabilities $\boldsymbol{y}^{\lambda}$ defined as:
\begin{equation}
    \hat{\boldsymbol{x}}_T^{\lambda} = D_{AE}(\boldsymbol{h}_T^{\lambda})
\end{equation}
\begin{equation}
    \hat{\boldsymbol{x}}_I^{\lambda} = D_{CAE}(\boldsymbol{h}_I^{\lambda})
\end{equation}
\begin{equation}
    \boldsymbol{y}^{\lambda} = C_{MLP}(\boldsymbol{h}^{\lambda})
\end{equation}
where $\boldsymbol{h}_T^{\lambda}$ and $\boldsymbol{h}_I^{\lambda}$ are given by:
\begin{equation}
    \boldsymbol{h}_T^{\lambda} = \boldsymbol{h}_T - \lambda \pdv{C_{MLP}(\boldsymbol{h})}{\boldsymbol{h}_T}
\end{equation}
\begin{equation}
    \boldsymbol{h}_I^{\lambda} = \boldsymbol{h}_I - \lambda \pdv{C_{MLP}(\boldsymbol{h})}{\boldsymbol{h}_I}
\end{equation}
so that $\boldsymbol{h}^{\lambda}$ is the concatenation of $\boldsymbol{h}_T^{\lambda}$ and $\boldsymbol{h}_I^{\lambda}$.

It is worth noting that finding an informative latent space  relies on the quality of the $AE$ and $CAE$.  
This justifies even more the use of the three-stage training, facilitating the training of the $AE$ and $CAE$.

\paragraph{Modality importance}

Since the multimodal embedding $\boldsymbol{h}$ is composed by the concatenation of the unimodal embeddings $\boldsymbol{h}_T$ and $\boldsymbol{h}_I$, we know to which modality each element of $\boldsymbol{h}$ is associated to.
Once calculated $\boldsymbol{h}_T^{\lambda}$ and $\boldsymbol{h}_I^{\lambda}$, we compute the modality normalized absolute differences, to understand how much each element has been shifted:
\begin{equation}
    \Delta_T = \frac{{||\boldsymbol{h}_T - \boldsymbol{h}_T^{\lambda}||_1}}{n} \label{eq:delta_T}
\end{equation}
\begin{equation}
    \Delta_I = \frac{{||\boldsymbol{h}_I - \boldsymbol{h}_I^{\lambda}||_1}}{m} \label{eq:delta_I}
\end{equation}
where $n$ and $m$ denote the number of elements in each vector, and $||.||_1$ denotes the $l_1$-$norm$.
Hence, $\Delta_T$ and $\Delta_I$ express the importance of each modality: the more a modality embedding has changed, the more important it is for the classification of a given sample.

\paragraph{Feature importance}

Similar to the modality importance, we now focus on an approach to reveal which features per modality are more important for the classification of a certain instance.
Using the shifted reconstructions $\hat{\boldsymbol{x}}_T^{\lambda}$ and $\hat{\boldsymbol{x}}_I^{\lambda}$, we compute the absolute differences with the original reconstructions $\hat{\boldsymbol{x}}_T$ and $\hat{\boldsymbol{x}}_I$
\begin{equation}
    \hat{\boldsymbol{\Delta}}_T = |\hat{\boldsymbol{x}}_T - \hat{\boldsymbol{x}}_T^{\lambda}| \label{eq:delta_T_tilde}
\end{equation}
\begin{equation}
    \hat{\boldsymbol{\Delta}}_I = |\hat{\boldsymbol{x}}_I - \hat{\boldsymbol{x}}_I^{\lambda}| \label{eq:delta_I_tilde}
\end{equation}
Note that $\hat{\boldsymbol{\Delta}}_T$ and $\hat{\boldsymbol{\Delta}}_I$ make us understand for each feature how much it has changed for the classification shift. 
The more a feature changes, the more important it is for the classification.
This works for both modalities, resulting $\hat{\boldsymbol{\Delta}_T}$ to be an importance vector for the tabular modality and $\hat{\boldsymbol{\Delta}_I}$ to be an importance matrix for the imaging modality.

\paragraph{Putting our method in the taxonomy}

Following the taxonomy introduced in section~\ref{sec:background} and originally presented in~\cite{bib:joshi2021review}, our proposal is an hybrid between during and post-hoc modelling as it exploits a specific network architecture using both perturbations and backpropagation methods to extract the explanations.
In particular our method uses counterfactual explanations, specifying the minimal desired changes required to flip the decision, mapping the class-specific and discriminative features of each modality.
In addition, our MXAI method is local, since we are interested in explaining how the model functions at instance level.
Finally, our method is model-specific since its architecture is constructed in a way to output the explanations.

\section{Experimental configuration} \label{sec:experimental}

In this section we introduce the used dataset and how the two modalities are pre-processed to train the network.
Then, we deepen the validation phase of the proposed MXAI method where we conducted a reader study with four COVID-19 expert radiologists.

\subsection{Dataset}

For the last two years the world has been struck by the COVID-19 pandemic causing millions of cases and deaths.
During this period, many researchers practitioners and companies have developed novel AI methods and tools to combat the rising of the pandemic by deepening the virus's understanding.
Many studies have focused their attention on unimodal data using CXR, CT or clinical examinations to replace or to supplement the reverse transcriptase-polymerase chain reaction tests.
But given the multimodal nature of medicine, both imaging data and clinical information can help radiologists and practitioners on determining the source of symptoms, stratifying the disease severity, and establishing the best treatment plan for the patient's specific needs.

We use the AIforCOVID imaging archive~\cite{bib:soda2021aiforcovid} because it is the only publicly available multimodal dataset on COVID-19 stratification, as shown in the survey~\cite{bib:santa2021public}.
The archive includes clinical data (tabular modality) and CXR scans (imaging modality) of 820 patients recorded from six different Italian hospitals.
In particular, there are 120, 104, 31, 139, 101, and 325 patients per hospital.
The interested readers can refer to~\cite{bib:soda2021aiforcovid,bib:guarrasi2022pareto,bib:guarrasi2022optimized,bib:guarrasi2021multi,bib:guarrasi2023multi} for further details.

The patients' data were collected at the time of hospitalization if the TR-PCR test resulted positive to the SARS-CoV-2 infection.
All the patients were assigned to the mild or severe class, on the basis of the clinical outcome.
The mild group includes  384 patients who were either sent back to domiciliary isolation or hospitalized without any ventilatory support, whereas  the severe group is composed of 436 patients who required non-invasive ventilation support, intensive care unit admission, or those who died.
Furthermore, any AI model trained on the AIforCOVID dataset is exposed to a diverse range of patient populations since it incorporates data from multiple centers, which should help ensure that the model is more generalizable and applicable to a wider extent. 


\subsection{Pre-processing}\label{subsec:pre-processing}

We applied the same pre-processing procedure and validation approach presented in~\cite{bib:soda2021aiforcovid} to avoid any performance bias, which are now briefly summarized for the sake of presentation.

\paragraph{Tabular data}
We use the 34 clinical descriptors indicated in~\cite{bib:soda2021aiforcovid} which are not direct indicators of the prognosis.
Missing data were imputed using the mean and the mode for continuous and categorical variables, respectively.
A min-max scaler was applied along the variables to have the features all in the same range $[0,1]$.

\paragraph{Imaging data}
This modality consists of CXR scans, which were processed by extracting the segmentation mask of lungs, using a U-Net trained on two non-COVID-19 lung datasets~\cite{bib:shiraishi2000development, bib:jaeger2014two}. 
The mask was used to extrapolate the minimum squared bounding box containing both lungs.
The extracted box was then resized to $224 \times 224$ matrix, and normalized with a min-max scaler bringing the pixel values in the range $[0,1]$.

\subsection{Implementation setup}

Here we describe the architectures of the three blocks of the model, as well as the parameters and settings used during training.

The AE's input and output layers consist of 34 (one for each feature) and 2 (one for each class) neurons, respectively.
Its encoder and decoder are composed of fully connected hidden layers activated by ReLU functions.
We opted to use such architectures since these feed-forward networks are able to learn a low-dimensional representation before being fused with the other modality~\cite{bib:glorot2010understanding}.
Both $E_{AE}$ and $D_{AE}$ have 6 hidden layers and $n=8$.
The loss function $L_{T}$ is the mean squared error (MSE).

The $CAE$ is a 2D ResNet101~\cite{bib:he2016deep}, which we selected because of the skip connections that mitigate the problem of the vanishing gradient, ensuring high fidelity image reconstruction.
Both the input and the output of the network have a $224 \times 224$ dimension, so that the dimension of the embedding is $m=4608$.
The dimensions the embeddings of both the $AE$ and the $CAE$ were chosen small enough to prevent the curse of dimensionality.
To facilitate the reconstruction training, this model was pre-trained trained on 4 different CXR datasets~\cite{bib:cohen2022torchxrayvision}, that in total account for a total of 674525 scans.
For consistency, the corresponding loss function $L_{I}$ is the MSE.

Let us now focus on the $C_{MLP}$ classifier: its input and output layers consist of 4616 and 2 neurons (one for each class), respectively.
It is composed of 7 fully connected hidden layers (with 512, 256, 128, 64, 32, 16, 8 neurons, respectively) activated by ReLU functions, with a Softmax activation in the output layer.
We design this structure to gradually learn the classification from the multimodal embedding.
The loss function $L_{C}$ is the cross-entropy.

For all the three stages of the training, introduced in section~\ref{subsec:three_stage}, we adopt the same training procedure of~\cite{bib:soda2021aiforcovid}, now summarized.
To prevent overfitting of the $CAE$, we applied the following image random transformations: horizontal or vertical shift ($-20 \leq$ pixels $\leq 20$), random zoom ($0.9 \leq$ factor $\leq 1.1$), vertical flip, random rotation ($-15^\circ \leq$ angle $\leq 15^\circ$), and elastic transform ($20 \leq \alpha \leq 40$, $\sigma=7$).
No augmentation was applied on the tabular data.
The loss functions $L_T$, $L_I$, $L_C$ and $L$ are regulated by an Adam optimizer with an initial learning rate of 0.001, which is scheduled to reduce by an order of magnitude every time the minimum validation loss does not change for 10 consecutive epochs.
To prevent overtraining and overfitting we fixed the number of maximum epochs to 300, with an early stopping of 25 epochs on the validation loss.

\subsection{Validation approach}

To understand the robustness of the model we trained the network in 10-fold stratified cross-validation (CV), and leave-one-center-out CV (LOCO), following the same experimental procedure described in~\cite{bib:soda2021aiforcovid}, thus ensuring a fair competition between the approaches.
In CV, the fold distribution of the training, validation and testing sets is 70\%-20\%-10\%, respectively.
In LOCO validation we study how the models generalize to different data sources, since in each fold the test set contains all the samples belonging to only one of the six hospitals that, of course, are not in the training and validation sets.

All the experiments were performed using a batch size of 16 on a NVIDIA TESLA A100 GPU with 32 GB of memory, using PyTorch as the deep learning library.

\subsection{Sanity check} \label{subsec:sanity}
To study the validity of the proposed MXAI method, we conducted a reader study with four radiologists assessing the prognosis of 96 patients randomly extracted.
Each radiologist has more than 10 years of experience.
The radiologists $R_1$, $R_2$, $R_3$, $R_4$ were presented with a survey that has two aims.
The first is to compare our method's classification performance with the one of human experts.
The second is to understand if the importance metrics $\Delta_T$, $\Delta_I$, $\hat{\boldsymbol{\Delta}}_T$ and $\hat{\boldsymbol{\Delta}}_I$, returned by our method, are coherent with the ones selected by the radiologists, which we denote as $\Delta_T^{R_i}$, $\Delta_I^{R_i}$, $\hat{\boldsymbol{\Delta}}_T^{R_i}$ and $\hat{\boldsymbol{\Delta}}_I^{R_i}$.
In particular, $\Delta_T^{R_i}$, $\Delta_I^{R_i}$ are the modality importance for the $i^{th}$ radiologist, and $\hat{\boldsymbol{\Delta}}_T^{R_i}$, $\hat{\boldsymbol{\Delta}}_I^{R_i}$ are the unimodal feature importance vector and matrix for the $i^{th}$ radiologist.
The survey was executed in double-blind, where no interaction between the radiologists was permitted.

In the survey, each radiologist observed both data modalities at the same time for each patient and performs the prognosis task.
Afterwards, the radiologists have to attribute an importance score, on a scale from 1 to 5, indicating how much significant each modality was for the prognosis task.
The grading of the scores are: 1 insignificant, 2 a bit significant, 3 neutral, 4 significant, 5 important.
A Softmax activation is applied constraining such values on the range $[0,1]$, where 0 means that the considered modality has no importance and 1 attributes the maximum importance.
As mentioned before, these modality importance are denoted as $\Delta_T^{R_i}$, $\Delta_I^{R_i}$.
The radiologist has the possibility to attribute the same importance to each modality if he/she believes that, for that patient, both modalities had the same impact in the decision.

Then, to understand the most important features for each modality, we asked the radiologist to select the clinical variables and to segment the areas of interest in the X-ray image most useful to stratify the patient, collecting $\hat{\boldsymbol{\Delta}}_T^{R_i}$ and $\hat{\boldsymbol{\Delta}}_I^{R_i}$, which are boolean, where elements equal to $1$ correspond to important features.
On both $\hat{\boldsymbol{\Delta}}_T^{R_i}$ and $\hat{\boldsymbol{\Delta}}_I^{R_i}$ a min-max normalization is applied on each instance, putting the elements on the range $[0,1]$, where 0 means that the considered feature has no importance and 1 attributes the max importance.

In the case of modality importance, we would expect a high intersection between the information reported by the radiologists $\Delta_T^{R_i}$ and $\Delta_I^{R_i}$ with the output of our method $\Delta_T$ and $\Delta_I$, respectively; the same holds in the case of unimodal feature importance, when comparing $\hat{\boldsymbol{\Delta}}_T^{R_i}$ and $\hat{\boldsymbol{\Delta}}_I^{R_i}$ with $\hat{\boldsymbol{\Delta}}_T$ and $\hat{\boldsymbol{\Delta}}_I$, respectively.


The surveys were executed on Google Forms, and the tool utilized to show and to segment the CXR scans was ITK-SNAP~\cite{bib:yushkevich2016itk}.

\subsection{Statistical analysis} \label{subsec:statistical_analysis}

The accuracy, the sensitivity and the specificity are the evaluation metrics used to assess the classification performance, as in~\cite{bib:soda2021aiforcovid}.
To asses if there exists a difference between the performance of our model and the baseline model we apply the one-way ANOVA and, to interpret the statistical significance, we used the pairwise Tukey test with a Bonferroni $p$-value correction at $\alpha=0.05$.

As described at the end of the previous section, we validate the MXAI modality importance $\Delta_T$, $\Delta_I$ and the feature importance $\hat{\boldsymbol{\Delta}}_T$, $\hat{\boldsymbol{\Delta}}_I$, comparing them with the importance proposed by the radiologists $\Delta_T^{R_i}$, $\Delta_I^{R_i}$ and $\hat{\boldsymbol{\Delta}}_T^{R_i}$, $\hat{\boldsymbol{\Delta}}_I^{R_i}$, respectively.

For the modality importance, we calculate the Pearson correlation $\rho$, and the paired sample t-test between the vector of importance modality ($\Delta_T$ and $\Delta_I$) over the instances given by our method, with the corresponding importance ($\Delta_T^{R_i}$ and $\Delta_I^{R_i}$) vector reported by the radiologists.
With the resulting statistics we can comprehend the measure of dependency between our method and the radiologists.
The higher $\rho$, the more our explanations are coherent with the importance scores reported by the radiologists, if the resulting t-test is not statistically significant ($p$-value $> 0.05$).

Turning our attention to the feature importance, we compute the intersection over union (IoU) between the feature importance proposed by the radiologists ($\hat{\boldsymbol{\Delta}}_T^{R_i}$ and $\hat{\boldsymbol{\Delta}}_I^{R_i}$) and the importance resulting from the latent shift ($\hat{\boldsymbol{\Delta}}_T$ and $\hat{\boldsymbol{\Delta}}_I$), respectively.
For the tabular modality $IoU_T$, we take the important features presented by the radiologists $\hat{\boldsymbol{\Delta}}_T^{R_i}$ and the binarized feature vector $\hat{\boldsymbol{\Delta}}_T^b$ (such that the values of $\hat{\boldsymbol{\Delta}}_T$  which are $< 0.5$ are set to $0$ and the ones $\geq 0.5$ are set to $1$), and compute
\begin{equation}
    IoU_T = \frac{\hat{\boldsymbol{\Delta}}_T^{R_i} \cap \hat{\boldsymbol{\Delta}}_T^b}{\hat{\boldsymbol{\Delta}}_T^{R_i} \cup \hat{\boldsymbol{\Delta}}_T^b}
\end{equation}
The higher the metric, the more concurrences there are between our method and the human annotation.
Similarly, when analyzing the imaging modality $IoU_I$ we take the segmented mask returned by the radiologists $\hat{\boldsymbol{\Delta}}_I^{R_i}$ and the binarized attribution map $\hat{\boldsymbol{\Delta}}_I^b$ (such that the values of $\hat{\boldsymbol{\Delta}}_I$  which are $< 0.5$ are set to $0$ and the ones $\geq 0.5$ are set to $1$), and compute
\begin{equation}
    IoU_I = \frac{\hat{\boldsymbol{\boldsymbol{\Delta}}}_I^{R_i} \cap \hat{\boldsymbol{\boldsymbol{\Delta}}}_I^b}{\hat{\boldsymbol{\boldsymbol{\Delta}}}_I^{R_i} \cup \hat{\boldsymbol{\boldsymbol{\Delta}}}_I^b}
\end{equation}
As before, the higher the metric, the more concurrences there are between our method and the human annotations.

\section{Results and discussion} \label{sec:results}

In this section we present the results obtained, dividing the discussion into five subsections that, in order,  deal with the classification and reconstruction performance, the three-stage training assessment, MXAI performance, an ablation study and forthcoming clinical  applications.  

\begin{table}[t]
\centering
\caption{Classification performance.}
\begin{adjustbox}{width=0.9\textwidth}
\begin{tabular}{lllll}
\toprule
\textbf{Model} & \textbf{Validation}   & \textbf{Accuracy} (\%) & \textbf{Sensitivity} (\%) & \textbf{Specificity} (\%) \\
\midrule
\multirow{3}{*}{\begin{tabular}[c]{@{}l@{}}Our proposal\\ (three-stage training)\end{tabular}}       & CV     & 76.75$\pm$5.32 & 78.58$\pm$6.48  & 74.55$\pm$5.86  \\
                            & LOCO   & 74.21$\pm$6.08 & 76.73$\pm$18.88 & 68.40$\pm$15.46 \\
                            & Survey & 76.77          & 78.54           & 74.57           \\
\midrule
\multirow{2}{*}{AIforCOVID~\cite{bib:soda2021aiforcovid}} & CV     & 76.90$\pm$5.40 & 78.80$\pm$6.40  & 74.70$\pm$5.90  \\
                            & LOCO   & 74.30$\pm$6.10 & 76.90$\pm$18.90 & 68.50$\pm$15.50 \\
\midrule
$R_1$                       & Survey & 68.75          & 43.75           & 93.75           \\
$R_2$                       & Survey & 72.92          & 70.83           & 75.00           \\
$R_3$                       & Survey & 76.04		  & 70.83           & 81.25           \\
$R_4$                       & Survey & 72.92          &  62.50          & 83.33           \\
\midrule
\multirow{3}{*}{\begin{tabular}[c]{@{}l@{}}Our proposal\\ (one-stage training)\end{tabular}}   & CV     & 70.38$\pm$1.78   & 72.57$\pm$1.72 & 68.62$\pm$1.12 \\
                            & LOCO   & 68.35$\pm$1.17   & 70.92$\pm$1.08 & 62.16$\pm$1.89 \\
                            & Survey & 70.48            & 72.51          & 68.67          \\
\bottomrule
\end{tabular}
\end{adjustbox}
\label{tab:classification}
\end{table}

\subsection{Classification and reconstruction performance}

\tablename~\ref{tab:classification} shows the classification performance and its columns specify: the learning model (human radiologists included), the validation approach (CV, LOCO or the reduced set of images used for the survey), the evaluation metrics employed, i.e., accuracy, sensitivity and specificity, for which we show the mean and the standard deviation of the metric in CV and in LOCO.
The table is organized into four horizontal sections: the first, the second and the fourth report the performance attained by learning models, whereas the third shows the performance of the four radiologists.
In particular, the first section shows results of our proposal and the second includes the best baseline model presented by~\cite{bib:soda2021aiforcovid}, which were attained by the hybrid approach.

Given that our method aims to increase the explainability of the model and not necessarily increase the performance, we first verify that with the co-learning we do not have a drop in performance with respect to the baseline~\cite{bib:soda2021aiforcovid}.
The results show that our model, even if it co-learns two tasks at once, obtains a performance similar to~\cite{bib:soda2021aiforcovid}.
In fact, the differences between our model and the baseline on all metrics, in both CV and LOCO, are not statistically significant ($p$-value $> 0.05$).
This suggests that our method is resilient to the notion that a decrease in performance is required to obtain better explanations.

We now compare our results with those of the radiologists: our proposal provides larger accuracy and sensitivity, while the specificity is lower.
This happens because predicting the prognosis of a patient affected by COVID-19 is a difficult task, giving a hint that AI could aid practitioners in the decision-making process.

\begin{table}[t]
\centering
\caption{Reconstruction performance.}
\begin{adjustbox}{width=0.6\textwidth}
\begin{tabular}{llll}
\toprule
\textbf{Model} & \textbf{Validation} & \textbf{Modality} & \textbf{MSE} \\
\midrule
\multirow{6}{*}{\begin{tabular}[c]{@{}l@{}}Our proposal\\ (three-stage training)\end{tabular}}        & CV     & $T$ &  0.04$\pm$0.01 \\
                            & LOCO   & $T$ &  0.05$\pm$0.02 \\
                            & Survey & $T$ &  0.04          \\ \cline{2-4}
                            & CV     & $I$ &  0.03$\pm$0.01 \\
                            & LOCO   & $I$ &  0.04$\pm$0.02 \\
                            & Survey & $I$ &  0.03          \\
\midrule
\multirow{6}{*}{\begin{tabular}[c]{@{}l@{}}Our proposal\\ (one-stage training)\end{tabular}}  & CV     & $T$ &  0.09$\pm$0.04 \\
                            & LOCO   & $T$ &  0.11$\pm$0.05 \\
                            & Survey & $T$ &  0.09          \\ \cline{2-4}
                            & CV     & $I$ &  0.07$\pm$0.02 \\
                            & LOCO   & $I$ &  0.09$\pm$0.03 \\
                            & Survey & $I$ &  0.07          \\
\bottomrule
\end{tabular}
\end{adjustbox}
\label{tab:reconstruction}
\end{table}

Let us recall that our method is jointly trained to classify and reconstruct the inputs via the autoencoders: for this reason in \tablename~\ref{tab:reconstruction} we show the MSE of $AE$ and $CAE$, i.e., the two autoencoders working with the $T$ and $I$ modalities, respectively.
As in \tablename~\ref{tab:classification}, here we have a similar row-column organization.
Turning our attention to the first section of this table, it is worth noting that the small values of the MSE confirm  the high quality of the reconstruction for both modalities.
This ensures that our MXAI method can provide good interpretability since it relies on the quality of such  reconstructions.

\subsection{Three-stage training assessment}

To validate the three-stage training introduced in section~\ref{subsec:three_stage}, we compare the classification and reconstruction performance with those attained adopting a one-stage training, which consists of skipping phases 1 and 2 of our method and directly train, in an end-to-end manner, the entire network with the combined loss $L$, without any pre-training.
The corresponding results are shown in the last section of Tables~\ref{tab:classification} and~\ref{tab:reconstruction}.
In the case of classification performance (\tablename~\ref{tab:classification}), we observe that the one-stage training provides lower performance than our three-stage proposal, whatever the performance metric and whatever the validation approach.
Furthermore, such performance differences are all statistically significant ($p$-value $< 0.01$).
Similar considerations hold in the case of reconstruction performance (\tablename~\ref{tab:reconstruction}).
These results confirm the usefulness of the three-stage training procedure, which aids the multimodal joint model in converging to a better solution.

\begin{table}[t]
\centering
\caption{$\rho$ (on the lower triangular) and the corresponding t-test $p$-value (upper triangular) of the modality importance, computed for each pair between our model and the radiologists.}
\begin{adjustbox}{width=0.6\textwidth}
\begin{tabular}{l|lllll}
\toprule
      & Our proposal & $R_1$ & $R_2$ & $R_3$ & $R_4$ \\
\midrule
Our proposal  & -    & 0.26 & 0.50 & 0.37 & 0.42 \\
$R_1$ & 0.78 & -    & 0.28 & 0.32 & 0.45 \\
$R_2$ & 0.84 & 0.90 & -    & 0.35 & 0.50 \\
$R_3$ & 0.77 & 0.82 & 0.78 & -    & 0.41 \\
$R_4$ & 0.79 & 0.77 & 0.83 & 0.85 & -    \\
\bottomrule
\end{tabular}
\end{adjustbox}
\label{tab:modality_importance}
\end{table}

\subsection{MXAI performance}

We now focus on validating the explanations provided by our proposal. Specifically, using the patients included in the survey, we compare the modality and feature explanations of our model to the importance reported by the radiologists.

Before going deep with the results, let us recall that in section~\ref{subsec:MXAI}, we formally put in relationship the counterfactual explanations with the data (equations \ref{eq:delta_T}, \ref{eq:delta_I}, \ref{eq:delta_T_tilde}, \ref{eq:delta_I_tilde}). 
Indeed, in the case of modality importance ($\Delta_T$ and $\Delta_I$), a counterfactual explanation highlights how large is the perturbation of the abstract representation of the clinical features or of the images caught by the latent space (equations~\ref{eq:delta_T} and~\ref{eq:delta_I}). 
In the case of the importance of each feature ($\hat{\boldsymbol{\Delta}}_T$ and $\hat{\boldsymbol{\Delta}}_I$), a counterfactual explanation works at level of each descriptor: for clinical data it represents how large is the variation between the original and the reconstructed clinical information (equation~\ref{eq:delta_T_tilde}), whereas for imaging data it measures pixels variations (equation~\ref{eq:delta_I_tilde}).  
The quantities defined in such four  equations are then considered in the sanity check (section~\ref{subsec:sanity}), which we introduced to validate the MXAI method.
 
According to section~\ref{subsec:statistical_analysis}, where we explain how we quantitatively compare the explanations provided by the model and those provided by the four radiologists, \tablename~\ref{tab:modality_importance} shows the Pearson correlation $\rho$ and the corresponding t-test $p$-values computed between the importance vector of a modality reported by our method and the importance vector reported by each radiologist.
These results reveal a high measure of dependency between our method and the radiologists and among the radiologists, suggesting that our model gives reasonable modality importance while producing the prognosis.

\begin{table}[t]
\centering
\caption{$IoU_T$ (lower triangular) and the $IoU_I$ (upper triangular) of the feature importance, computed for each pair between our model and the radiologists.}
\begin{adjustbox}{width=0.9\textwidth}
\begin{tabular}{l|lllll}
\toprule
      & Our proposal           & $R_1$          & $R_2$          & $R_3$          & $R_4$           \\
\midrule
Our proposal  & -              & 59.62$\pm$3.13 & 62.97$\pm$2.61 & 63.77$\pm$2.25 & 64.63$\pm$2.76  \\
$R_1$ & 52.37$\pm$3.12 & -              & 60.56$\pm$2.78 & 63.42$\pm$3.39 & 61.81$\pm$3.28  \\
$R_2$ & 54.72$\pm$2.86 & 52.37$\pm$2.78 & -              & 60.98$\pm$3.76 & 62.44$\pm$2.99  \\
$R_3$ & 53.52$\pm$3.21 & 54.69$\pm$3.42 & 51.23$\pm$2.98 & -              & 63.36$\pm$3.64  \\
$R_4$ & 51.31$\pm$2.69 & 52.73$\pm$3.31 & 54.66$\pm$2.79 & 55.43$\pm$3.05 & -               \\
\bottomrule
\end{tabular}
\end{adjustbox}
\label{tab:feature_importance}
\end{table}

Let us now consider the unimodal feature importance: in this case \tablename~\ref{tab:feature_importance} shows the $IoU_T$ and $IoU_I$ for each possible pair between our model and the radiologists.
As mentioned in section~\ref{subsec:statistical_analysis}, these metrics permit us understand the coherence between the returned feature importance.
These scores not only show that the radiologists have a fairly high degree of intersection of important features among each other, but also that the degree of intersection is of the same magnitude even with our proposal.
This implies that our model concentrates on the relevant features of each modality when making the decision on the prognosis.

As stated in section~\ref{sec:background}, there is currently a lack of multimodal XAI methods in the literature. 
Therefore, to further validate the performance of our explanations, we compared the unimodal explanations generated by our method with other well-established XAI methods, namely Integrated Gradients, LIME, and SHAP~\cite{bib:arrieta2020explainable}.
We selected these methods because they can be applied to both tabular and imaging modalities, they are model-agnostic, i.e., they can be used with any model irrespective of its underlying architecture, and they all offer local explanations.
Specifically, we extract the explanations by utilizing the $C_{MLP}$, $E_{AE}$, and $E_{CAE}$ modules for each modality, respectively. 
In \tablename~\ref{tab:unimodal_competitors}, we present the average $IoU_T$ and $IoU_I$ across the feature importance scores from all the radiologists for both our proposal and the competing methods.
The results demonstrate that our unimodal explanations are not statistically different from these XAI methods ($p$-value $> 0.05$), indicating that our approach is coherent with state-of-the-art methods from a unimodal perspective.
Furthermore, it is worth noting that our proposed method not only introduces unimodal explanations but also incorporates multimodal explanations, a novel feature that is not available in existing XAI techniques.

\begin{table}[t]
\centering
\caption{Comparison of $IoU_T$ and $IoU_I$ of the feature importance, computed between XAI methods and the radiologists. The values in the table are the mean and standard deviation of $IoU$ across all radiologists.}
\begin{adjustbox}{width=0.55\textwidth}
\begin{tabular}{l|ll}
\toprule
\textbf{XAI Method} & $IoU_T$ & $IoU_I$ \\
\midrule
Our Proposal         & 52.98$\pm$2.97 & 62.75$\pm$2.69 \\
Integrated Gradients & 53.10$\pm$3.05 & 63.20$\pm$2.83 \\
LIME                 & 52.70$\pm$3.22 & 62.90$\pm$2.98 \\
SHAP                 & 53.05$\pm$3.12 & 63.10$\pm$2.91 \\
\bottomrule
\end{tabular}
\end{adjustbox}
\label{tab:unimodal_competitors}
\end{table}

\begin{table}[t]
\centering
\caption{Classification performance in the ablation study}
\begin{adjustbox}{width=0.95\textwidth}
\begin{tabular}{llllll}
\toprule
\textbf{Model} & \textbf{Validation} & \textbf{Modality} & \textbf{Accuracy} (\%) & \textbf{Sensitivity} (\%) & \textbf{Specificity} (\%) \\
\midrule
\multirow{6}{*}{Our Proposal}       & CV     & $T$ & 75.78$\pm$0.75 & 76.63$\pm$0.71  & 74.74$\pm$1.02  \\
                            & LOCO   & $T$ & 73.48$\pm$3.20 & 69.87$\pm$3.11  & 79.56$\pm$8.60  \\
                            & Survey & $T$ & 75.69          & 76.65           & 74.71           \\ \cline{2-6}
                            & CV     & $I$ & 74.14$\pm$1.03 & 74.57$\pm$1.78  & 73.96$\pm$1.21  \\
                            & LOCO   & $I$ & 70.46$\pm$1.06 & 72.03$\pm$1.02  & 69.55$\pm$1.61  \\
                            & Survey & $I$ & 74.32          & 74.42           & 73.88           \\
\midrule
\multirow{4}{*}{AIforCOVID~\cite{bib:soda2021aiforcovid}} & CV     & $T$ & 75.70$\pm$0.80 & 76.00$\pm$0.70  & 75.40$\pm$1.10  \\
                            & LOCO   & $T$ & 73.40$\pm$4.40 & 69.90$\pm$15.80 & 79.50$\pm$13.60 \\ \cline{2-6}
                            & CV     & $I$ & 74.20$\pm$1.00 & 74.80$\pm$1.90  & 73.80$\pm$1.30  \\
                            & LOCO   & $I$ & 70.50$\pm$1.00 & 72.00$\pm$1.10  & 69.60$\pm$1.50  \\
\bottomrule
\end{tabular}
\end{adjustbox}
\label{tab:classification_ablation}
\end{table}

\begin{table}[t]
\centering
\caption{Reconstruction performance of the ablation study.}
\begin{adjustbox}{width=0.5\textwidth}
\begin{tabular}{llll}
\toprule
\textbf{Model} & \textbf{Data} & \textbf{Modality}  & \textbf{MSE} \\
\midrule
\multirow{6}{*}{Our Proposal}       & CV     & $T$ &  0.03$\pm$0.01 \\
                            & LOCO   & $T$ &  0.04$\pm$0.02 \\
                            & Survey & $T$ &  0.03          \\ \cline{2-4}
                            & CV     & $I$ &  0.02$\pm$0.01 \\
                            & LOCO   & $I$ &  0.03$\pm$0.02 \\
                            & Survey & $I$ &  0.02          \\
\bottomrule
\end{tabular}
\end{adjustbox}
\label{tab:reconstruction_ablation}
\end{table}

\begin{table}[t] 
\centering
\caption{$IoU_T$ (lower triangular) and the $IoU_I$ (upper triangular) of the feature importance for models trained in ablation, computed for each pair between our model and the radiologists.}
\begin{adjustbox}{width=0.9\textwidth}
\begin{tabular}{l|lllll}
\toprule
      & Our proposal           & $R_1$          & $R_2$          & $R_3$          & $R_4$ \\
\midrule
Our proposal  & -              & 60.31$\pm$2.98 & 62.45$\pm$2.31 & 63.57$\pm$2.49 & 64.22$\pm$3.58  \\
$R_1$ & 53.63$\pm$2.45 & -              & 60.56$\pm$2.20 & 62.10$\pm$3.10 & 61.90$\pm$2.98  \\
$R_2$ & 54.65$\pm$2.95 & 52.45$\pm$3.21 & -              & 61.07$\pm$3.30 & 62.90$\pm$3.20  \\
$R_3$ & 54.49$\pm$3.03 & 53.99$\pm$3.32 & 50.89$\pm$3.00 & -              & 62.88$\pm$3.50  \\
$R_4$ & 50.40$\pm$3.01 & 52.41$\pm$3.40 & 52.54$\pm$3.11 & 55.30$\pm$2.87 & -               \\
\bottomrule
\end{tabular}
\end{adjustbox}
\label{tab:modality_importance_ablation}
\end{table}

\subsection{Ablation study}
We now discuss what happens when only one modality is available.
To this end, we ran two experiments: in the first we removed the $AE$ and we trained again the other part of the model, i.e., we worked only with the imaging modality disregarding the tabular clinical data.
In the second we flipped the ablation, removing the $CAE$ and, thus, we did not consider the images.
\tablename~\ref{tab:classification_ablation} shows the results we achieved in the case of experiments ran in CV, LOCO and using the survey images.
Furthermore, the third columns specifies the modality used.
As before, we also show the hybrid baseline model presented in~\cite{bib:soda2021aiforcovid}, which is trained only on one modality.
The results of our proposal shows that using only the tabular clinical data or only the imaging data provides similar results, which do not statistically differ from each other, whatever the performance score considered ($p$-value $> 0.05$).
Furthermore, in comparison with the performance of the full multimodal approach (first section of \tablename~\ref{tab:classification}), we notice that both unimodal models report a statistically significant drop in performance ($p$-value $< 0.01$), whatever the validation approach or the performance score.
As before, we notice that our model, even if it co-learns two tasks at once, obtains similar performance to the hybrid model, i.e., the best baseline model of~\cite{bib:soda2021aiforcovid}.
In fact, the difference between our model and the baseline model on all metrics, in both CV and LOCO, is not statistically significant ($p$-value $> 0.05$).

For completeness, \tablename~\ref{tab:reconstruction_ablation} shows the reconstruction results in terms of MSE for each modality.
In comparison with \tablename~\ref{tab:reconstruction}, as expected, we notice that the reconstruction error has significantly decreased ($p$-value $> 0.05$) becaause the model can focus on one modality at a time, making it easier to learn an efficient embedding mapping.

As before, we also investigate the results in terms of explainability.
Straightforwardly, in this case the modality importance does not make sense, so that \tablename~\ref{tab:modality_importance_ablation} reports the $IoU_T$ and the $IoU_I$ of new unimodal feature importance.
These results show that, even if we have a drop in classification and reconstruction performance, the explanations are consistent between the radiologists and between our method and the radiologists, suggesting that the MXAI method is robust to a missing modality~\cite{bib:rofena2024deep,bib:caruso2024deep}.

\begin{figure}[tp]
\centering
\caption{Four case studies: for each we show the feature importance indicated by our proposal ($\hat{\boldsymbol{\boldsymbol{\Delta}}}_T$ and $\hat{\boldsymbol{\boldsymbol{\Delta}}}_I$) and the corresponding important features indicated by radiologists ($\hat{\boldsymbol{\boldsymbol{\Delta}}}_T^{R_i}$ and $\hat{\boldsymbol{\boldsymbol{\Delta}}}_I^{R_i}$) for the tabular and the imaging modalities, respectively. The rows show examples of patients with mild (first and third row) and severe (second and fourth row) outcomes, for both success (first and second row) and failure cases (third and fourth row) of our model.}
\includegraphics[width=\textwidth]{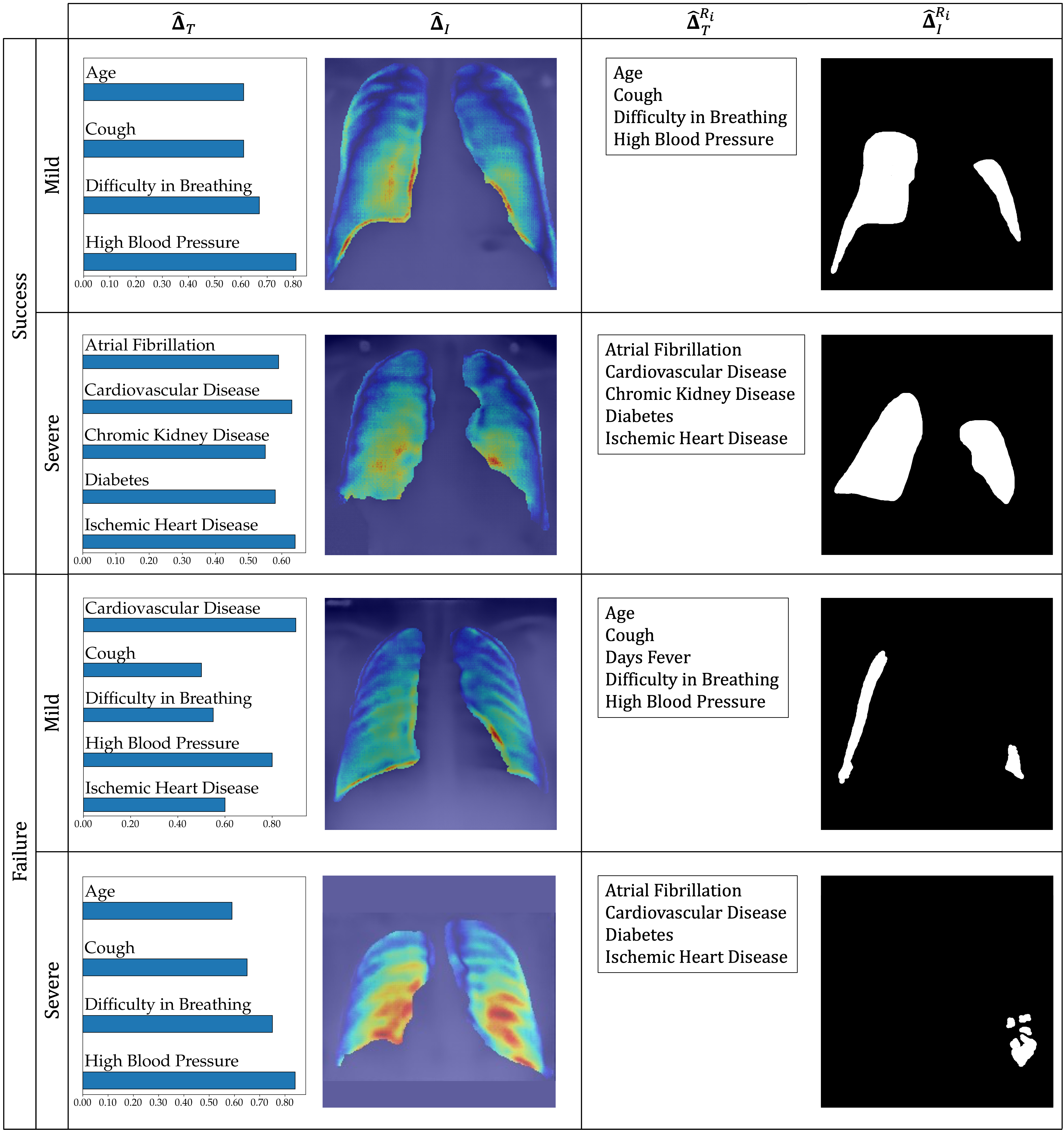}
\label{fig:xai_example}
\end{figure}

\subsection{Clinical perspective and case studies}

In a clinical practice scenario, we believe that our AI system can serve as a precursor to subsequent multimodal research for predicting the evolution of COVID-19. Its classifications and explanations can assist radiologists in performing prognosis tasks.

Indeed, on the one side,  in~\cite{bib:stephanie2020determinants} the authors showed that  the rise of X-ray severity over the course of COVID-19 infection  increases  the sensitivity of COVID-19 detection using CXR over time ($55\%$ at $\leq 2$ days to $79\%$ at $> 11$ days), whilst it decreases the specificity ($83\%$ at $\leq 2$ days to $70\%$ at $> 11$ days).
On the other side, as~\tablename~\ref{tab:classification} shows, our proposal  provides  a   larger sensitivity than the radiologists, suggesting that it  can anticipate the evolution of positive COVID-19 cases, that is in an initial phase of the disease when the patient accesses the emergency department, our approach achieves a sensitivity ($78.56\%$) equal to that which the X-ray alone shows after several days.

Furthermore, the proposed architecture has the beneficial feature to offer transparent decisions since, for each patient, the radiologists can observe at the same time the original data (both the clinical features and the X-ray image), the modality importance $\Delta_T$ and $\Delta_I$, and the unimodal feature importance $\hat{\boldsymbol{\Delta}}_T$ and $\hat{\boldsymbol{\Delta}}_I$.
With $\Delta_T$ and $\Delta_I$, the radiologists would be guided to understand on which modality to concentrate more on.
Instead, with $\hat{\boldsymbol{\Delta}}_T$ and $\hat{\boldsymbol{\Delta}}_I$ we guide the radiologist to concentrate on certain clinical characteristics and on specific areas of the X-ray scan.

Figure~\ref{fig:xai_example} presents four case studies.
It is organized in four columns: the first two show the feature importance indicated by our proposal ($\hat{\boldsymbol{\boldsymbol{\Delta}}}_T$ and $\hat{\boldsymbol{\boldsymbol{\Delta}}}_I$), whereas the third and the fourth show the corresponding important features indicated by radiologists ($\hat{\boldsymbol{\boldsymbol{\Delta}}}_T^{R_i}$ and $\hat{\boldsymbol{\boldsymbol{\Delta}}}_I^{R_i}$), for the tabular and the imaging modalities, respectively.
The tabular clinical data importance $\hat{\boldsymbol{\Delta}}_T$ is represented as a bar-plot on the $\hat{\boldsymbol{\Delta}}_T$ column, so that the longer the bar, the more important the clinical variable is.
For the sake of visualization, we only show the features part of $\hat{\boldsymbol{\Delta}}_T^b$, keeping the magnitude of the importance computed according to equation~\ref{eq:delta_T_tilde}.
The X-ray image importance map $\hat{\boldsymbol{\Delta}}_I$ is shown as a heatmap on the original scan, which represents the relevance of each pixel in the image for the prognosis task on a color scale ranging from blue (low importance) to red (high importance) on the $\hat{\boldsymbol{\Delta}}_I$ column.
Looking at the figure by row, the first group of rows shows two success cases, where our classifier correctly classifies the patient's outcome, whereas the second group of rows shows two failure cases, where our classifier incorrectly classifies the patient's outcome.
In both cases of classification success or failure, we present an example from both the mild and severe classes.
Note that all these cases have been correctly classified by the radiologist.
In the success examples, we note a strong agreement between our proposal and the radiologist for both modalities. 
Specifically, in both mild and severe cases, the tabular features are the same, and 
the most important pixels for the model largely overlap with the area of interest segmented by the radiologist. 
In the failure cases, our model assigns importance to tabular features and to image regions that are different from those highlighted by the radiologist. 
We speculate that this may be the reason for the model incorrect predictions.
In particular, in the mild case, only three out of five most important tabular features coincide with those suggested by the human expert, whilst  for the severe case there is no  overlap. 
Additionally, if we turn our attention to the images, both  the mild and severe cases exhibited a very low intersection between the  most important  regions for the models and the manually segmented areas.


\section{Conclusion} \label{sec:conclusion}

In this work we presented an end-to-end multimodal architecture that jointly learns modality reconstructions and multimodal classification using tabular clinical and imaging data.
With respect to the literature using such modalities for medical tasks, we deem that our method is the only one which offers intrinsic model-specific local multimodal explanations.
In particular, multimodal explanations are computed by exploiting the latent space learnt by jointly training the end-to-end architecture and using a latent shift-based counterfactual method. 
We tested our approach in the context of the COVID-19 pandemic using the AIforCOVID public dataset, which includes both X-ray and clinical data. 
The extensive quantitative experimentation shows that the latent space retains features useful to succeed both in a reconstruction and classification task and, thus, resulting in an informative space for the latent-shift method.
Moreover, the sanity check, although very time-consuming, was very useful since it showed a high intersection between the explanations provided by the method and those of the radiologists, both for the modality and the feature importance.

A   reflection on this work highlights two main limitations. 
The first is that the reliability of the explanations the method produces  is constrained by its reliance on the classification and reconstruction performance of the model.
As all components of the method are data-driven and there is no dedicated module for explainability, the generalizability of the explanations could be limited by the quality of the data.
In this respect, we plan to evaluate the effectiveness of our methodology on different datasets from different domains and with different data types.
The second limitation stems from noticing that, although our proposal identifies the importance of each modality and the importance of each unimodal feature per sample, it does not 
find out high-level concepts, including those expert-based.   
To cope with this issue we deem that concept knowledge mining~\cite{bib:qi2020towards,bib:graziani2020concept} could be a viable solution that we plan to investigate in future work to enable human experts to better understand how the prediction is identified by the model.

\section*{Acknowledgements}

The computations of this work were enabled by resources provided by the Swedish National Infrastructure for Computing (SNIC), partially funded by the Swedish Research Council through grant agreement no. 2018-05973.
This work was partially supported by: i) “University-Industry Educational Centre in Advanced Biomedical 411 and Medical Informatics (CEBMI)” (Grant agreement no. 612462-EPP-1-2019-1-SK-EPPKA2-KA, 412 Educational, Audiovisual and Culture Executive Agency of the European Union);
ii) PNRR MUR project PE0000013-FAIR;
iii) FONDO PER LA CRESCITA SOSTENIBILE (F.C.S.) Bando Accordo Innovazione DM 24/5/2017 (Ministero delle Imprese e del Made in Italy), under the project entitled ``Piattaforma per la Medicina di Precisione. Intelligenza Artificiale e Diagnostica Clinica Integrata'' (CUP B89J23000580005).

\section*{Author Contributions}

V.G.: Conceptualization, Methodology, Software, Validation, Formal analysis, Investigation, Data Curation, Writing - Original Draft, Writing - Review \& Editing, Visualization.
L.T.: Conceptualization, Methodology, Software, Validation, Formal analysis, Data Curation.
D.A.: Validation.
E.F.: Validation.
D.F.: Validation.
D.S.: Validation.
P.S.: Conceptualization, Methodology, Validation, Formal analysis, Writing - Review \& Editing, Supervision, Funding acquisition.

\bibliographystyle{elsarticle-num} 
\bibliography{bib.bib}

\end{document}